\documentclass[letterpaper, 10 pt, conference]{ieeeconf}     % Use this line for a4 paper

\IEEEoverridecommandlockouts                              % This command is only needed if 
\usepackage{lipsum}  
\usepackage{amsmath,amssymb,amsfonts}
\usepackage{amssymb}
\usepackage{algorithmic}
\usepackage{graphicx}
\usepackage{textcomp}
\usepackage{color}
\usepackage{xcolor}
\usepackage{url}
\usepackage{bm}
\usepackage[hidelinks]{hyperref}
\usepackage{acronym}
\usepackage{siunitx}
\usepackage{subcaption}

\usepackage{booktabs}

\usepackage[textsize=footnotesize]{todonotes}

\newcommand{\Rthree}[0]{\mathbb{R}^3}

\usepackage{tikz}
\usepackage{pgfplots}
\pgfplotsset{compat=1.14}
\usetikzlibrary{backgrounds,arrows,automata,shapes,positioning,calc,through,spy,shapes,shapes.geometric,shapes.multipart,fit,patterns,fadings}
\pgfdeclarelayer{background}
\pgfdeclarelayer{foreground}
\pgfsetlayers{background, main, foreground}

\title{\LARGE \bf NMPP: Nonlinear Model Predictive Planning for\\Agile UAV Flight in Cluttered Environments}

\author{Ondřej Procházka, Juraj Pauko, Robert Pěnička, Martin Saska% <-this % stops a space
\thanks{The authors are with the Multi-robot Systems Group, Faculty of Electrical
Engineering, Czech Technical University in Prague, Czech Republic (\protect\url{http://mrs.felk.cvut.cz/}). 
This work has been supported by the Czech Science Foundation (GAČR) under research project No. 23-06162M, by the European Union under the project Robotics 
and Advanced Industrial Production (reg. no. CZ.02.01.01/00/22\_008/0004590) 
and by CTU grant no SGS26/077/OHK3/1T/13.
(Ondřej Procházka and Juraj Pauko contributed equally to this work.)%% <-this % stops a space
}}

\begin{document}

% must precede every other citation; configures IEEEtran.bst (see ref.bib)
\bstctlcite{BSTcontrol}

% UAV states
\newcommand{\position}{p}
\newcommand{\positionvector}{\bm{\position}}
\newcommand{\quaternion}{q}
\newcommand{\orientation}{\bm{\quaternion}}
\newcommand{\linearvel}{v}
\newcommand{\linearvelvector}{\bm{\linearvel}}
\newcommand{\angularvel}{\omega}
\newcommand{\angularvelvector}{\bm{\angularvel}}
\newcommand{\integralerror}{z}
\newcommand{\integralerrorvector}{\bm{\integralerror}}
\newcommand{\statederivative}{\dot{\statestd}}
\newcommand{\statestd}{\bm{x}}
\newcommand{\uavstate}{\statestd}
\newcommand{\statepid}{\bm{x}}

% UAV inputs
\newcommand{\commandedangularvel}{\angularvel_c}
\newcommand{\propvel}{\Omega}
\newcommand{\propvelvector}{\bm{\propvel}}
\newcommand{\propvelcmd}{\propvel_c}
\newcommand{\propvelcmdvector}{\bm{\propvelcmd}}
\newcommand{\dragforcevector}{\bm{f}_{D}}
\newcommand{\commandedtorquevector}{\bm{\tau_c}}
\newcommand{\torquevector}{\bm{\tau}}
\newcommand{\uavinput}{\bm{u}}
% thrusts and its vectors
\newcommand{\motorforce}{f}
\newcommand{\thrustvector}{\bm{f}}
\newcommand{\commandedthrustvector}{\bm{f_c}}
\newcommand{\normalizedrpm}{r}
\newcommand{\normalizedrpmvector}{\bm{\normalizedrpm}}
\newcommand{\throttle}{t}
\newcommand{\throttlevec}{\bm{\throttle}}
\newcommand{\commandedcollectivethrottle}{\throttle_c}
\newcommand{\collectivethrust}{T}
\newcommand{\commandedcollectivethrust}{\collectivethrust_c}
\newcommand{\bodythrustvector}{\bm{f}_{T}}
% Others
\newcommand{\motorallocationmat}{\bm{M}}
\newcommand{\mixermat}{\bm{G}}
\newcommand{\inertiamat}{\bm{J}}
\newcommand{\mpchorizon}{N}
\newcommand{\errpenmat}{\mathbf{Q}}
\newcommand{\inputpenmat}{\mathbf{R}}
\newcommand{\terminalpenmat}{\mathbf{P}}
\newcommand{\weightmat}{\mathbf{W}}                  % nonlinear-LS weight
\newcommand{\thrustpenmat}{\mathbf{R}_{\thrustvector}}       % thrust-state penalty
\newcommand{\angaccpenmat}{\mathbf{Q}_{\dot{\angularvel}}}   % angular-acceleration penalty
\newcommand{\costres}{\bm{y}}                        % nonlinear-LS residual
\newcommand{\slackvec}{\bm{s}}                       % slack variables
\newcommand{\errormat}{\mathbf{\Tilde{\statestd}}}
\newcommand{\errorinputmat}{\mathbf{\Tilde{\uavinput}}}
\newcommand{\statestddesired}{\overset{*}{\statestd}}
\newcommand{\taumot}{k_{mot}}
\newcommand{\thrustcoef}{c_f}

\newcommand{\acceleration}{a}
\newcommand{\accelerationvector}{\bm{\acceleration}}
\newcommand{\jerk}{j}
\newcommand{\jerkvector}{\bm{\jerk}}
\newcommand{\basisthree}{\bm{e}_3}

% --- cross-reference names used by \autoref{} ---
\renewcommand{\figureautorefname}{Fig.}
\renewcommand{\tableautorefname}{Table}
\renewcommand{\equationautorefname}{Eq.}
\renewcommand{\sectionautorefname}{Section}
\renewcommand{\subsectionautorefname}{\sectionautorefname}
\renewcommand{\subsubsectionautorefname}{\sectionautorefname}

\newcommand{\reffig}[1]{Fig.~\ref{#1}}
\newcommand{\refeq}[1]{Eq.~\ref{#1}}
\newcommand{\reftable}[1]{Table~\ref{#1}}
\newcommand{\medialink}{https://youtu.be/pWIOeak4yeE}

\acrodef{pid}[PID]{Proportional Integral Derivative}
\acrodef{uav}[UAV]{Unmanned Aerial Vehicle}
\acrodef{vtol}[VTOL]{Vertical Take-Off and Landing}
\acrodef{mpc}[MPC]{Model Predictive Controller}
\acrodef{nmpc}[NMPC]{Nonlinear Model Predictive Control}
\acrodef{nmpp}[NMPP]{Nonlinear Model Predictive Planning}
\acrodef{dfbc}[DFBC]{Differential-Flatness-Based Controller}
\acrodef{indi}[INDI]{Incremental Nonlinear Dynamic Inversion}
\acrodef{ocp}[OCP]{Optimal Control Problem}
\acrodef{imu}[IMU]{Inertial Measurement Unit}
\acrodef{rti}[RTI]{Real-Time Iteration}
\acrodef{rpm}[RPM]{Rotations per minute}
\acrodef{rmse}[RMSE]{Root Mean Square Error}
\acrodef{qp}[QP]{Quadratic Programming}
\acrodef{gnss}[GNSS]{Global Navigation Satellite System}
\acrodef{lidar}[LiDAR]{Light Detection and Ranging}
\acrodef{rtk}[RTK]{Real-time Kinematics}
\acrodef{gmppi}[GMPPI]{Geometric Model Predictive Path Integral}
\acrodef{pca}[PCA]{Principal Component Analysis}
\acrodef{cbf}[CBF]{Control Barrier Function}
\acrodef{hocbf}[HOCBF]{Higher-Order Control Barrier Function}
\acrodef{roi}[ROI]{Range Of Interest}
\acrodef{apf}[APF]{Artificial Potential Fields}
\acrodef{esdf}[ESDF]{Euclidean Signed Distance Field}
\maketitle
\thispagestyle{empty}
\pagestyle{empty}

\begin{abstract}
    Flying a quadrotor through a cluttered environment requires not only planning a collision-free reference trajectory based on perceived obstacles, but the reference also needs to be dynamically feasible and within the actuation limits of the vehicle, so that the controller can track it precisely.
    Existing methods either optimize a smooth polynomial inside a convex corridor, which limits agility, or treat obstacles as soft costs traded against tracking performance.
    We propose a \ac{nmpp} that imposes perceived obstacles as hard geometric constraints and hands a full-state reference to an obstacle-blind $SE(3)$ controller.
    % Warm-starting each optimization from the previous prediction keeps that constraint feasible in real time, while a fixed set of ellipsoids bounds the problem size.
    Our planner achieves a 58--\SI{67}{\percent} lower position RMSE than a linear Model Predictive Control trajectory planner and a 41--\SI{70}{\percent} lower RMSE than a polynomial trajectory planner.
    It also completes all forest flights with up to \SI{9.5}{\metre\per\second} speed without collisions, and achieves \SI{86}{\percent} flight success rate under a more aggressive speed profile where a state-of-the-art planner has only \SI{26}{\percent} success rate.
    The real-world deployment showed reliable execution flying up to \SI{5.5}{\metre\per\second} in an unknown cluttered environment.
    % with \SI{13.5}{\milli\second} per planning run on board.
\end{abstract}

\vspace{-0.5em}
\section*{Supplementary Material}
 {\footnotesize
  \vspace{-0.3em}
  \noindent \textbf{Video:} \href{\medialink}{\medialink} %\url{https://youtu.be/wTucpUq1Z80}
  \vspace{-0.7em}
 }

\section{Introduction}
% Why is it important?
% \begin{itemize}
%     \item decreasing the mismatch between the planned and the flown trajectory
%     \item separating the two stages fails in both directions: a collision-free polynomial path may not be dynamically feasible, and a dynamically feasible trajectory may not be collision-free
%     \item a controller that lags behind or cuts corners can collide even when the planned trajectory was collision-free
% \end{itemize}
Multirotor \acp{uav} are increasingly deployed in missions that require flight through cluttered and initially unknown environments, including search and rescue \cite{Doherty_2007_SAR} and infrastructure inspections \cite{Burri_2012_plant_inspection}.
In such missions, the achievable flight speed determines how much of the environment can be covered within the vehicle's limited endurance, which motivates operating the platform close to its dynamic limits.
Flying fast between obstacles requires a reference trajectory that is not only collision-free but also respects the \ac{uav}'s dynamics and actuation limits.
The obstacles are perceived by the onboard sensors only during the flight, so the trajectory has to be planned online.
The trajectory is often generated with simplified vehicle dynamics, causing a mismatch between the planned and flown trajectory \cite{Teissing_2024_PMM_mintime}.
Moreover, collision avoidance and dynamic feasibility are commonly handled in separate stages, where a collision-free path is first found in a map and then converted into a trajectory for the controller.
Thus, the path ignores the dynamics and the trajectory ignores the obstacles.

% Why is the problem hard? What makes it challenging?
% \begin{itemize}
%     \item The free space is nonconvex, so hard obstacle constraints make the NLP nonconvex and prone to local minima and infeasibility
%     \item It has to run onboard in real time using the current sensory data
%     \item The sensor returns a variable and unbounded number of obstacles, while the optimization problem has to keep a fixed size to give a predictable computation time
%     \item Even a node-feasible discrete solution can collide between shooting nodes
% \end{itemize}
This paper addresses the generation of such a trajectory in a single onboard optimization, which respects the \ac{uav} dynamics and actuation limits while remaining collision-free with respect to the obstacles perceived by the onboard \ac{lidar} sensor.
The planner relies only on onboard sensing and has to return a solution within a few milliseconds of each planning period.
The result is continuously handed to an $SE(3)$ geometric controller \cite{lee2010geometric} as a full-state reference, so that the controller can follow it without deviating into the obstacles it was planned around.

\begin{figure}[t]
	\centering
	\input{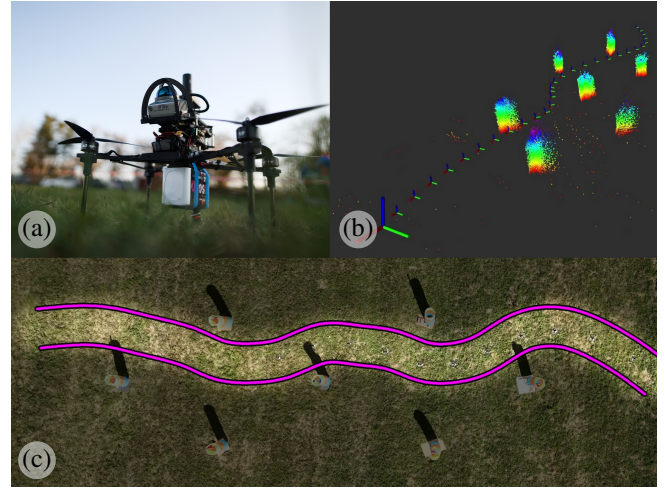}
    \vspace{-1.5em}
	\caption{Illustrative images from real-world experiments: (a) UAV used in the experiments, (b) flown path visualization in RViz, and (c) top view of the course with the flown path.\label{fig:intro_picture}}
	\vspace{-1.5em}
\end{figure}

% Why is the problem hard? What makes it challenging?
% \begin{itemize}
%     \item The free space is nonconvex, so hard obstacle constraints make the NLP nonconvex and prone to local minima and infeasibility
%     \item It has to run onboard in real time using the current sensory data
%     \item The sensor returns a variable and unbounded number of obstacles, while the optimization problem has to keep a fixed size to give a predictable computation time
%     \item Even a node-feasible discrete solution can collide between shooting nodes
% \end{itemize}
% How far has existing work come? What is the frontier?
% \begin{itemize}
%     \item Corridor-based polynomial optimization, which is efficient enough to run online but limited to smooth motions \cite{wang2022geometrically, ren2025safety}
%     \item Avoidance placed inside the control loop, which reacts to what is currently perceived but simplifies the vehicle dynamics or expresses the obstacles only as soft costs \cite{liang2025pointcloud, harms2025safe, pochobradsky2026geometric}
% \end{itemize}
Meeting these requirements in one optimization is difficult for three reasons.
First, the free space around obstacles is nonconvex, so imposing the obstacles as hard constraints makes the problem nonconvex and prone to local minima and infeasibility.
Second, the sensor returns a variable and unbounded number of obstacles, while the problem has to keep a fixed size for its computation time to stay predictable.
Third, a solution that is feasible at the discrete prediction nodes can still intersect an obstacle between two of them.
Existing work addresses these difficulties in two ways.
Corridor-based polynomial optimization restricts the trajectory to a convex subset of the free space, which removes the nonconvexity and allows online computation \cite{wang2022geometrically, ren2025safety}.
However, the smooth polynomial representation keeps the vehicle from flying close to its dynamic limits.
Placing the avoidance inside the control loop instead preserves the reaction to what is currently perceived \cite{liang2025pointcloud, harms2025safe, pochobradsky2026geometric}, but the obstacles are then expressed as soft costs, 
% , or the vehicle dynamics are simplified so that the problem remains solvable at the rate of the control loop
so that the avoidance is traded against tracking rather than guaranteed.

% Why hasn't the problem been solved? What is the stumbling block?
% \begin{itemize}
%     \item Hard obstacle constraints in a control-loop NMPC are fragile, the initial state is a noisy estimate that can land inside (or with no feasible escape from) the constrained set, making the QP infeasible in the loop, so real-time NMPC controllers relax to soft constraints.
%     \item Dense sensor data cannot enter a fixed-size optimization problem directly, so the environment has to be reduced to a compact representation without discarding what the trajectory still has to avoid
% \end{itemize}
% Keeping the full model and imposing the obstacles exactly is what the control loop makes difficult.
% A controller is initialized from the current state estimate, which is noisy and can fall inside the constrained set, or leave no feasible escape from it, so that the problem becomes infeasible at the rate at which the controller runs and the obstacle constraints have to be relaxed to keep it solvable.
% The perceived environment is a second difficulty, since dense sensor data cannot enter a fixed-size problem directly and has to be reduced to a compact representation without discarding what the trajectory still has to avoid.

% What does our paper contribute? what is the method we propose?
% What is the key idea? What is the magic trick? What is the new insight or technique that enables us to advance the frontier?
To this end, we propose \acf{nmpp}, which uses \ac{nmpc} as a trajectory planner that generates a full-state reference for a separate controller rather than actuator commands.
Each optimization is initialized from the previously generated prediction rather than from the noisy state estimate, which allows imposing the obstacle constraint exactly.
The accumulated point cloud is converted into ellipsoids, of which a fixed number enters the problem, keeping its size constant regardless of the scene.
The hard constraint is complemented by a soft obstacle penalty and a softened acceleration-level \ac{hocbf} condition \cite{xiao2022hocbf}.
The full obstacle set validates the trajectory between prediction nodes.
To enforce actuation limits efficiently, the \ac{nmpc} uses the thrust rate as its control input.
The thrust and angular-acceleration limits thus become linear constraints.
To the best of our knowledge, this is the first planner that imposes \ac{lidar}-perceived obstacles as hard constraints on the full quadrotor dynamics with thrust dynamics and runs onboard.

% What do the experiments say?
In simulation, \ac{nmpp} is compared with two trajectory planners, one based on linear \ac{mpc} \cite{baca2018model} and one based on polynomial trajectories \cite{wang2021large}.
All three provide their reference to the same unmodified $SE(3)$ controller \cite{lee2010geometric}, so the reference is the only component that differs.
Across five constraint profiles with speed limits, \ac{nmpp} achieves a 58--\SI{67}{\percent} lower position \ac{rmse} than the linear \ac{mpc} trajectory planner and a 41--\SI{70}{\percent} lower \ac{rmse} than the polynomial trajectory planner.
% Four constraint profiles (slow, medium, fast, super fast) are flown, with ten repetitions of each.
The obstacle avoidance is evaluated in a Poisson forest against SUPER~\cite{ren2025safety}, with both planners flying in the same world towards the same goal.
\ac{nmpp} completed all flights without collision at speeds up to \SI{9.5}{\metre\per\second} (\textit{super fast} profile), whereas the success rate of SUPER dropped to \SI{70}{\percent} under that profile.
Under a more aggressive profile, the two planners reached the goal in \SI{86}{\percent} and \SI{26}{\percent} of fifty flights, respectively.
Last but not least, we deployed the planner on a real quadrotor and validated it in a cluttered outdoor environment, as illustrated in \autoref{fig:intro_picture}. 
The \ac{uav} successfully avoided obstacles while flying at speeds up to \SI{5.5}{\metre\per\second}.

\section{Related Work}
\label{sec:sota}

Methods for autonomous \ac{uav} navigation in cluttered environments range from reactive collision avoidance to global, optimization-based, and learning-based trajectory planning.

Reactive methods compute avoidance commands directly from the current perception data. 
\ac{apf} represent obstacles through repulsive potentials \cite{khatib1985realtime}, while Velocity Obstacles identify control velocities that may lead to a future collision \cite{fiorini1998motion}. 
Extensions such as ORCA support reciprocal collision avoidance between multiple agents \cite{vanDenBerg2011reciprocal}.
Reciprocal velocity constraints have also been integrated into \ac{nmpc} for mutual collision avoidance in agile \ac{uav} flight \cite{kratky2025rvc}.
These approaches are computationally efficient, but their local decisions can lead to local minima and usually rely on simplified vehicle dynamics. 

Global motion planners instead search for a collision-free path in a representation of the environment. 
Graph-search and sampling-based algorithms, such as A*, RRT, and RRT*, can find paths through complex obstacle configurations \cite{karaman2011sampling}. 
The resulting path is commonly used to further plan a smooth trajectory that can be tracked by the \ac{uav}.
Minimum jerk/snap polynomial trajectories are widely used for this purpose because the differential flatness of a quadrotor allows smooth position trajectories to be mapped to reference attitude and thrust \cite{mellinger2011minimum, richter2016polynomial}. 
Efficient polynomial representations further enable the generation of long trajectories with many segments \cite{wang2021large}. 
Minimum-time formulations based on a point-mass model exploit the maneuverability of the \ac{uav} more effectively than polynomial trajectories, at the cost of a reference that is only partially feasible for the real vehicle~\cite{Teissing_2024_PMM_mintime}. 
However, these methods generally require predefined waypoints or a collision-free corridor and do not themselves determine how to avoid newly perceived obstacles. 

Optimization-based planners incorporate smoothness, dynamic feasibility, and collision avoidance into a single trajectory-planning problem. 
Corridor-based methods such as GCOPTER~\cite{wang2022geometrically} efficiently optimize polynomial trajectories while enforcing smoothness-based formulations. 
Receding safe corridors can also be constructed incrementally along the flight, so that high-speed smooth trajectories are planned within them \cite{ren2022bubble}. 
The corridor, however, restricts the trajectory to a convex subset of the free space, so the motion that can be produced is limited by how the corridor was carved rather than by the capability of the vehicle.

\ac{mpc} provides an alternative in which the future vehicle motion is optimized over a receding horizon. 
It naturally supports state, input, and actuation constraints and has therefore been applied to \ac{uav} trajectory tracking and collision avoidance \cite{baca2018model}. 
An \ac{nmpc} controller that integrates the dynamics of the low-level flight controller was shown to reduce the tracking error on agile references \cite{gupta2025lol}. 
There, the optimization is placed at the controller level, where it produces the commands for the actuators, whereas the method proposed here places it at the planner level, where it produces a reference for a separate controller. 
The CIAO and CIAO* methods, presented in \cite{schoels2020nmpc, schoels2020ciao}, respectively, formulate collision avoidance through convex inner approximations of free space, providing recursively feasible predictive motion planning in static and predictable dynamic environments. 
The principal difficulty is that a nonlinear geometric constraint has to be introduced for every obstacle that the optimization accounts for, so the size of the problem grows with the number of obstacles that are taken into account. 
When the environment is measured online, that number is neither known in advance nor bounded, which is what the following approaches, and \ac{nmpp} have to address. 
% \ac{lidar} sensors provide accurate three-dimensional obstacle measurements over a substantially longer range than commonly used depth cameras. 
% However, the raw point cloud cannot usually be introduced directly into a real-time trajectory optimization problem because it may contain thousands of points and its size changes between measurements. 
% Consequently, the perception system must transform the measurements into a representation suitable for collision queries and planning. 

Several planners therefore reduce or avoid dense volumetric mapping. 
NanoMap stores a local history of depth measurements and performs uncertainty-aware proximity queries without fusing all measurements into a global voxel-map~\cite{florence2018nanomap}. 
EGO-Planner removes the need for an \ac{esdf} by extracting obstacle information only when the optimized trajectory intersects an occupied region \cite{zhou2021egoplanner}. 
SUPER generates polyhedral flight corridors directly from \ac{lidar} point clouds and optimizes an exploratory and a backup minimum-snap polynomial trajectory within them, to distinguish high-speed progress from motion guaranteed to remain in observed free space \cite{ren2025safety}. 
These methods substantially reduce mapping overhead, but collision avoidance is still performed through path search, corridor construction, and polynomial trajectory optimization. 

A different line of work removes the modular pipeline altogether and learns a mapping from sensor data directly to control commands, which avoids the latency accumulated across mapping, planning and control \cite{loquercio2021learning, zhang2025learning}. 
These methods embed little prior knowledge about the controlled vehicle, so a change of the platform or its parameters requires retraining on large amounts of data. 

Another group of methods integrates perception directly into predictive control. 
Local point clouds can be queried directly using KD-trees to construct obstacle costs inside a point-mass \ac{mpc} formulation \cite{zhang2025mapless}. 
Point-cloud constraints can be incorporated into \ac{cbf}-based \ac{mpc} by selecting risk points close to the predicted motion \cite{liang2025pointcloud}. 
For quadrotors, multiple obstacle constraints can be aggregated into a composite \ac{cbf} and enforced by a high-rate safety filter operating on a third-order dynamic model~\cite{harms2025safe}. 
Alternatively, \ac{gmppi} approach sampled \ac{uav} rollouts directly into depth images~\cite{pochobradsky2026geometric} and evaluates collision without constructing an explicit geometric map. 
These approaches demonstrate the benefits of connecting local perception with predictive control, but they either simplify the vehicle dynamics, apply collision avoidance as a separate safety filter, or represent obstacles only through soft collision costs.
In contrast, \ac{nmpp} imposes the \ac{lidar}-perceived obstacles as hard constraints on the full quadrotor dynamics with thrust dynamics, without restricting the trajectory to a convex corridor.
A fixed number of ellipsoids keeps the problem size constant regardless of the scene, and the planner runs onboard while providing a full-state reference.

\section{Problem Formulation}
We consider the problem of online trajectory planning for a quadrotor flying toward one or a sequence of reference waypoints in an initially unknown environment. 
At each replanning instant, the objective is to generate a finite-horizon trajectory that progresses toward the active reference while respecting the nonlinear quadrotor dynamics and the physical limitations of the \ac{uav}.
In addition, the generated trajectory must remain collision-free with respect to obstacles observed by the onboard \ac{lidar} sensor.

% The trajectory-planning problem is formulated progressively. 
% First, the quadrotor dynamics and the finite-horizon \ac{nmpc} objective function are introduced. 
% The formulation is then extended with the actuation and motion constraints to ensure that the generated reference is dynamically feasible. 
% Finally, the locally perceived environment is incorporated through ellipsoidal obstacle constraints and trajectory validation. 

\subsection{UAV Model Definition}
Let the state be $\statestd=(\positionvector,\orientation,\linearvelvector,\angularvelvector)$, with position $\positionvector\in\mathbb{R}^3$, unit-quaternion attitude $\orientation$, linear velocity $\linearvelvector$ and body rate $\angularvelvector$, and let $\thrustvector\in\mathbb{R}^{4}$ be the vector of per-rotor thrusts $\motorforce_i$.
The rotors act on the body through the constant allocation
\begin{equation}\label{eq:motorallocationmat}
    \bodythrustvector = \collectivethrust\,\basisthree,
    \qquad
    \collectivethrust = \sum_{i=0}^{3}\motorforce_i,
    \qquad
    \torquevector     = \motorallocationmat\,\thrustvector, 
\end{equation}
\begin{equation}
    \motorallocationmat =
    \begin{bmatrix}
       y_0     &  y_1     &  y_2    &  y_3    \\
      -x_0     & -x_1     & -x_2    & -x_3    \\
      -\kappa  & -\kappa  &  \kappa &  \kappa
    \end{bmatrix},
\end{equation}
where $\bodythrustvector$ is the thrust vector in the body frame, $\collectivethrust$ is the collective thrust, $\motorforce_i$ is the thrust produced by the $i$-th rotor, $\torquevector$ is the torque produced in the body frame, and $\basisthree=(0,0,1)^\top$.
The allocation matrix $\motorallocationmat$ is determined by the rotor positions $(x_j,y_j)$ in the body frame and the rotor torque constant $\kappa$.
%totally unnecessary fluff
% The body frame has $x$ pointing forward and $y$ to the left, and the rotors are numbered so
% that $0$ and $3$ lie ahead of the centre of mass and $1$ and $2$ behind it, which makes
% $\{0,1\}$ and $\{2,3\}$ the two diagonal pairs of opposite spin direction.

The quadrotor dynamics are given by
\begin{subequations}
  \label{eq:quad_dyn}
  \begin{align}
      \bm{\dot{\position}}         & = \linearvelvector \text{,} \vphantom{\frac{1}{2}}
\label{eq:quad_dyn_dot_p}                                                 \\
      \bm{\dot{\linearvel}}        & =
\frac{\bm{R}(\orientation)(\bodythrustvector+\dragforcevector)}{m} +
\bm{g}\text{,}\label{eq:quad_dyn_dot_v}               \\
      \bm{\dot{\quaternion}}       & = \frac{1}{2} \orientation \circledast
\begin{bmatrix} 0 \\ \angularvelvector \end{bmatrix}\text{,} \label{eq:quad_dyn_dot_q}
                         \\
      \bm{\dot{\angularvelvector}} & = \inertiamat^{-1} (\torquevector -
\angularvelvector \times \inertiamat \angularvelvector)\text{,}\label{eq:quad_dyn_dot_w}
  \end{align}
\end{subequations}
where the operator $\circledast$ denotes the quaternion multiplication, $\bm{R}(\orientation)$ is the rotation matrix corresponding to the quaternion $\orientation$, $m$ is the mass of the \ac{uav}, $\bm{g}$ is the gravitational acceleration vector, and $\inertiamat$ is the diagonal inertia matrix of the rigid body.
The drag force is modelled as $\dragforcevector=-\bm{K}_v\bm{R}(\orientation)^\top\linearvelvector$, where $\bm{K}_v=\operatorname{diag}(k_{vx},k_{vy},k_{vz})$ contains the linear drag coefficients.

\subsection{Nonlinear Model Predictive Control Formulation}
The reference trajectory is generated by repeatedly solving a finite-horizon \ac{ocp}, stated in the form used by the \texttt{acados} framework \cite{Verschueren2021}.
Over a horizon of $\mpchorizon$ shooting nodes with time steps $\Delta t_k$, and in a real-time-iteration scheme, the planner solves
\begin{subequations}
    \label{eq:ocp}
    \begin{align}
        \min_{\statestd_{0:\mpchorizon},\,\uavinput_{0:\mpchorizon-1},\,\slackvec_{0:\mpchorizon}}
          \;&\sum_{k=0}^{\mpchorizon-1}\big(\ell(\statestd_k,\uavinput_k)+\Phi_s(\slackvec_k)\big)
          \nonumber\\
          &+\ell_e(\statestd_{\mpchorizon})+\Phi^{e}_s(\slackvec_{\mpchorizon})
          \label{eq:ocp_cost}\\
        \text{s.t.}\quad
          &\statestd_0=\hat{\statestd},\label{eq:ocp_init}\\
          &\statestd_{k+1}=f_{RK4}(\statestd_k,\uavinput_k,\Delta t_k),\label{eq:ocp_dyn}\\
          &\underline{\bm{h}}\le\bm{h}(\statestd_k,\uavinput_k)\le\overline{\bm{h}},\label{eq:ocp_hard}\\
          &\underline{\bm{h}}_s-\slackvec_k\le\bm{h}_s(\statestd_k,\uavinput_k)\le\overline{\bm{h}}_s+\slackvec_k,
          \label{eq:ocp_soft}\\
          &\slackvec_k\ge\bm{0},\label{eq:ocp_slack}
    \end{align}
\end{subequations}
where $k$ indexes the shooting nodes, $\uavinput_k$ is the control input, $\hat{\statestd}$ is the current state estimate, $\ell$ and $\ell_e$ are the stage and terminal costs, and $f_{RK4}$ denotes the discrete dynamics obtained by integrating \eqref{eq:quad_dyn} over the time step $\Delta t_k$ with a fourth-order Runge--Kutta scheme.
The vector $\bm{h}$ collects the constraints that are enforced exactly, and $\bm{h}_s$ those that are softened with the slack variables $\slackvec_k$, which are penalised by $\Phi_s(\slackvec)=\tfrac{1}{2}\slackvec^\top\bm{Z}\slackvec+\bm{z}^\top\slackvec$ with the quadratic and linear penalty weights $\bm{Z}$ and $\bm{z}$.
A subset of the softened constraints is imposed at the terminal node as well, with its own slack variables $\slackvec_{\mpchorizon}$ and penalty $\Phi^{e}_s$.
The stage cost, the two constraint sets, and the choice of the time steps $\Delta t_k$ are described in \autoref{sec:proposed_method}.

\section{Proposed Method}
\label{sec:proposed_method}
\begin{figure}
    \centering
    \includegraphics[width=1\columnwidth]{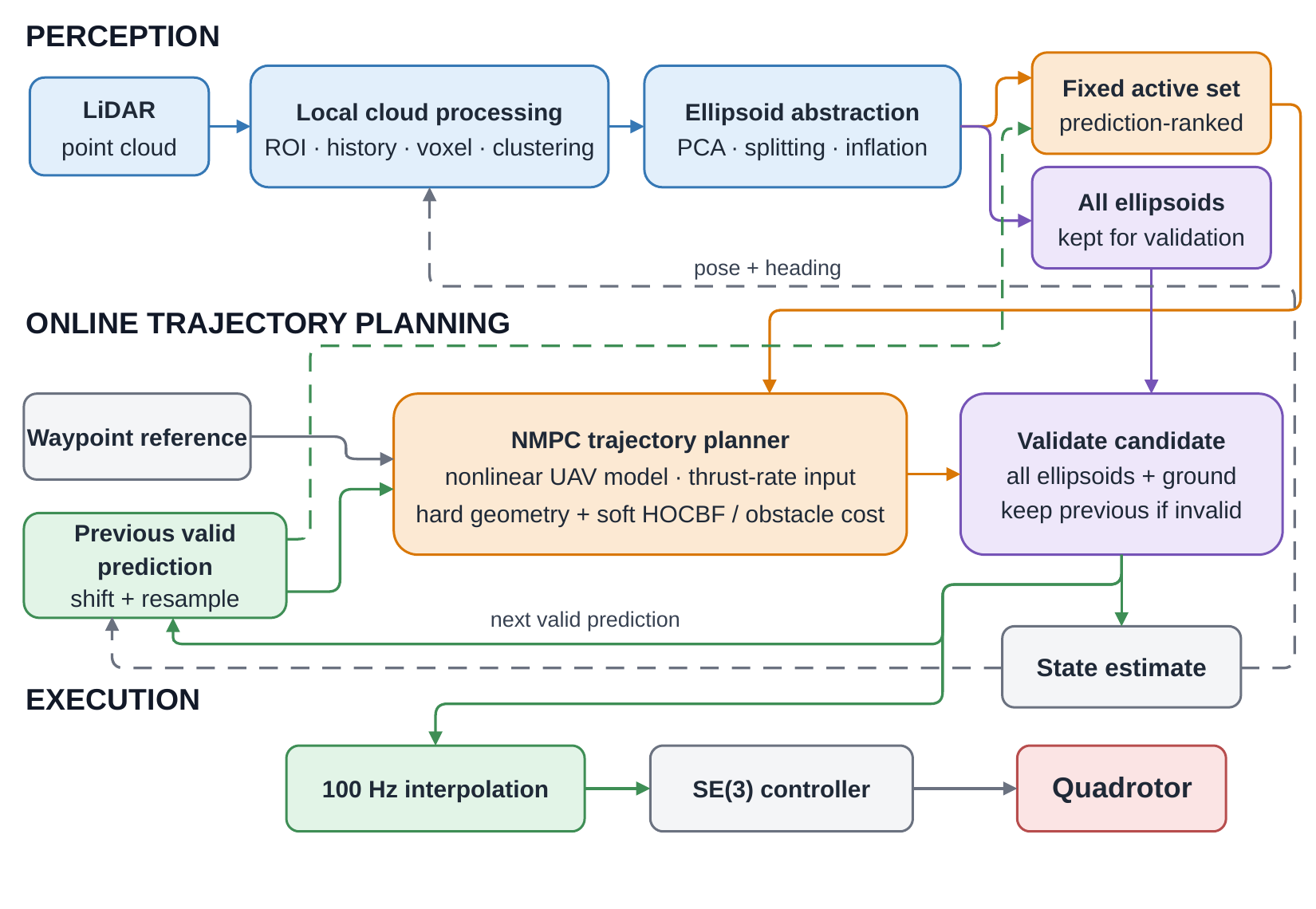}
    \caption{Whole \ac{nmpp} system pipeline.}
    \label{fig:pipeline}
    \vspace{-1.5em}
\end{figure}

In this section, we describe \ac{nmpp}, an \ac{nmpc} configuration that enables its use as a trajectory planner rather than a controller. 
The complete pipeline is summarized in \autoref{fig:pipeline}.

\subsection{Augmented NMPC Model}
\label{sec:augmented_model}
In the conventional formulation \cite{gupta2025lol}, the thrust is the control, $\uavinput=\thrustvector$.
We instead augment the state with the thrust and take its rate as the control,
\begin{equation}\label{eq:augmentation}
    \statestd=(\positionvector,\orientation,\linearvelvector,\angularvelvector,\thrustvector),
    \qquad
    \uavinput=\dot{\thrustvector}\in\mathbb{R}^4,
\end{equation}
leaving the dynamics \eqref{eq:quad_dyn} unchanged.

We use the thrust rate as the control because the solver keeps the control constant between two shooting nodes.
If the thrust itself is the control, it jumps at every node, and so do the linear and angular accelerations of the planned trajectory, which is then only $\positionvector\in C^1$ and $\angularvelvector\in C^0$.
Writing the linear acceleration as $\accelerationvector=\dot{\linearvelvector}$ and differentiating \eqref{eq:quad_dyn_dot_v} and \eqref{eq:quad_dyn_dot_w}, using $\dot{\bm{R}}=\bm{R}[\angularvelvector]_\times$ and the allocation \eqref{eq:motorallocationmat}, gives
\begin{subequations}
    \label{eq:jerk_all}
    \begin{align}
      \begin{split}
        \jerkvector = \dot{\accelerationvector}
          &= \frac{\bm{R}(\orientation)\,[\angularvelvector]_\times
             \big(\bodythrustvector+\dragforcevector\big)}{m}\\
          &\quad + \frac{\bm{R}(\orientation)\big(\big(\sum_{i=0}^{3}\dot{\motorforce}_i\big)
             \basisthree + \dot{\dragforcevector}\big)}{m}\text{,}
      \end{split} \label{eq:jerk}                                  \\
      \ddot{\angularvelvector}       & =
    \inertiamat^{-1}\!\big(\motorallocationmat\,\dot{\thrustvector}
      - \dot{\angularvelvector}\times\inertiamat\angularvelvector
      - \angularvelvector\times\inertiamat\dot{\angularvelvector}\big)\text{,}
    \label{eq:angjerk}
    \end{align}
\end{subequations}
where both the jerk and the angular jerk depend linearly on~$\dot{\thrustvector}$.
With the thrust rate as the control, the thrust changes continuously, the accelerations no longer jump, and the reference becomes one order smoother, $\positionvector\in C^2$ and $\angularvelvector\in C^1$.

\subsection{Constraints}
\label{sec:constraints}
  With the thrust as a state and its rate as the control, the actuator limits become linear bounds on the state and the control.
  The bounds
  \begin{subequations}
      \label{eq:constraints}
      \begin{align}
          % \uavinput_k                                          & \in
  % \big[-\dot{\motorforce}_{max},\,\dot{\motorforce}_{max}\big]\text{,}
                 % & & \text{rotor slew rate}        \label{eq:con_slew} \\
          \thrustvector_k                                      & \in
  [\motorforce_{min},\,\motorforce_{max}]\text{,}
                 & & \text{per-rotor thrust}       \label{eq:con_thrust} \\
          \collectivethrust_k                                  & \in
  [\collectivethrust_{min},\,\collectivethrust_{max}]\text{,}
                 & & \text{collective thrust}      \label{eq:con_coll} \\
          \inertiamat^{-1}\motorallocationmat\,\thrustvector_k  & \in
  \big[-\dot{\angularvelvector}_{max},\,\dot{\angularvelvector}_{max}\big]\text{,}
                & & \text{ang. acceleration}   \label{eq:con_angacc}
      \end{align}
  \end{subequations}
  form the hard constraints \eqref{eq:ocp_hard}, while
  \begin{subequations}
      \label{eq:constraints_soft}
      \begin{align}
          \angularvelvector_k                                  & \in
  [\angularvelvector_{min},\angularvelvector_{max}]\text{,}
                 & & \text{body rate}              \label{eq:con_rate} \\
          \linearvelvector_k                                   & \in
  [\linearvelvector_{min},\linearvelvector_{max}]\text{,}
                 & & \text{linear velocity}        \label{eq:con_vel} \\
          \big[\bm{R}(\orientation_k)\big]_{33}                & \ge \cos\theta_{max}\text{,}
                 & & \text{tilt}                   \label{eq:con_tilt} \\
          \big|\bm{n}_i^\top\bm{\quaternion}_{xy,k}\big|       & \le
  \sin\tfrac{\theta_{max}}{2}\text{,}
                 & & \text{tilt polygon}           \label{eq:con_oct}
      \end{align}
  \end{subequations}
  are implemented as softened constraints using the slack variables in \eqref{eq:ocp_soft}. 
  This permits temporary violations of the motion limtis, for example after switching to a more restrictive constraint profile, without making the optimization infeasible. 
  The actuator constraints \eqref{eq:con_thrust}-\eqref{eq:con_angacc}, however remain hard. 
  In \eqref{eq:con_oct}, $\bm{\quaternion}_{xy,k}=\begin{bmatrix}\quaternion_{x,k} & \quaternion_{y,k}\end{bmatrix}^\top$ contains the $x$ and $y$ components of the attitude quaternion $\quaternion_k$. 
  The vectors $\mathbf{n}_i$, $i = 1, \ldots, 4$, define the directions used to construct the octagonal approximation of the circular tilt region in the $(\quaternion_x, \quaternion_y)$ plane, with $n_1 = (1, 0)\top$, $n_2 = (0, 1)\top$ and $n_{3,4} = \frac{1}{\sqrt{2}}(1, \pm1)\top$. 
  The limits on the linear acceleration are not imposed on the acceleration itself, but through the collective thrust \eqref{eq:con_coll} and the tilt \eqref{eq:con_tilt}--\eqref{eq:con_oct}.
  Neglecting drag, the vertical and horizontal accelerations of the \ac{uav} are $a_z = \frac{\collectivethrust}{m}\cos\theta - g$ and $a_{xy} = \frac{\collectivethrust}{m}\sin\theta$, where $\theta$ is the tilt angle and $g$ is the gravitational acceleration.
  The bounds on the vertical acceleration are therefore enforced by tightening the collective thrust limits to
  \begin{equation}
      \label{eq:acc_thrust}
      \begin{split}
          \collectivethrust_k \in \big[
            &\max\big(\collectivethrust_{min},\,m(g - a^{-}_{z,max})\big),\\
            &\min\big(\collectivethrust_{max},\,m(g + a^{+}_{z,max})\big)\big]\text{,}
      \end{split}
  \end{equation}
  and the bound on the horizontal acceleration by replacing the tilt limit with
  \begin{equation}
      \label{eq:acc_tilt}
      \theta_{max} \leftarrow \min\Big(\theta_{max},\,
        \arcsin\frac{a_{xy,max}}{g + a^{+}_{z,max}}\Big)\text{,}
  \end{equation}
  which is the largest tilt at which the upper thrust limit of \eqref{eq:acc_thrust} still produces at most $a_{xy,max}$.
  Since $[\bm{R}(\orientation)]_{33}=1-2(\quaternion_x^2+\quaternion_y^2)$, the tilt constraint \eqref{eq:con_tilt} is equivalent to $\quaternion_x^2+\quaternion_y^2 \le \sin^2\frac{\theta_{max}}{2}$, and \eqref{eq:con_oct} is a linear outer approximation of this disc.
  Using both constraints allows the solver to bound the tilt in the linear general-constraint block of the \ac{qp}, while \eqref{eq:con_tilt} retains the exact condition.
  For the same reason, the gyroscopic contribution $\angularvelvector\times\inertiamat\angularvelvector$ of \eqref{eq:quad_dyn_dot_w} is neglected in \eqref{eq:con_angacc}, so that the bound stays linear in the state instead of becoming a nonlinear path constraint.
  It is small relative to $\motorallocationmat\thrustvector$ in the operating regime considered.

\subsection{Collision Avoidance}
\label{sec:collision_avoidance}
The collision-avoidance method connects local \ac{lidar} perception directly with the \ac{nmpc} trajectory planner. 
The main difficulty is that a point cloud contains thousands of points and an unpredictable number of obstacles, whereas the \ac{nmpc} must retain a fixed problem size to provide predictable onboard computation. 
Therefore, the measured environment is first converted into a compact set of ellipsoidal obstacles. 
A fixed number of the most relevant obstacles is included in the optimization, while the complete perceived obstacle set is retained for subsequent trajectory validation. 

\subsubsection{Point-Cloud Processing}
Let $\mathcal{P} = \{\mathbf{\xi}_1, \ldots, \mathbf{\xi}_{N_p}\}, \quad \mathbf{\xi}_i \in \mathbb{R}^3$ denote an accumulated LiDAR point cloud expressed in the static planning frame. 
Points corresponding to the \ac{uav} body and ground are removed, and the cloud is restricted to a local heading-aligned \ac{roi} around the vehicle. 
The onboard Livox MID-360 uses a non-repetitive scanning pattern.
To obtain a more complete representation of obstacle surfaces, the consecutive point clouds are accumulated over non-overlapping \(0.20\,\mathrm{s}\) windows to obtain denser obstacle coverage.

The cloud $\mathcal{P}$ is downsampled using a range-adaptive voxel grid \cite{rusu2011pcl}. 
A fine resolution is retained near the \ac{uav}, where accurate obstacle boundaries are important, while increasingly coarse resolutions are used farther away. 
Euclidean clustering then partitions the processed cloud into 
\begin{equation}
    \mathcal{C} = \{ \mathcal{C}_1, \ldots, \mathcal{C}_{N_c} \}, \qquad \mathcal{C}_m \subseteq \mathcal{P},
\end{equation}
where $\mathcal{C}_m$ denotes an obstacle cluster.

\subsubsection{Obstacle Representation}
Each cluster $\mathcal{C}_j$ is represented by one or more oriented ellipsoids. 
Compared with a sphere, an ellipsoid provides a tighter approximation of elongated structures such as tree trunks, poles, and walls.
\ac{pca} determines the orientation of the ellipsoid, after which its center and radii are selected such that all points in $\mathcal{C}_j$ are enclosed. 
An additional inflation safety margin accounts for the vehicle size in the perceived obstacle boundary.

The resulting perceived obstacle set is denoted by
\begin{equation}
    \mathcal{E} = \{ \mathcal{E}_1, \ldots, \mathcal{E}_{N_e} \}.
\end{equation}
An ellipsoid $\mathcal{E}_j$ with center $\mathbf{c}_j$ is represented by
\begin{equation}
    \mathcal{E}_j = \left\{ \mathbf{\xi}\in\mathbb{R}^3 \,\middle|\, (\mathbf{\xi}-\mathbf{c}_j)^\top \mathbf{A}_j (\mathbf{\xi}-\mathbf{c}_j) \leq 1 \right\},
\end{equation}
where the positive-definite matrix $\mathbf{A}_j$ contains its orientation and radii.
Equivalently, the obstacle function is defined as
\begin{equation}
    h_j(\mathbf{p}) = (\mathbf{p} - \mathbf{c}_j)^\top \mathbf{A}_j (\mathbf{p} - \mathbf{c}_j) - 1.
    \label{eq:obstacle_function}
\end{equation}
A point lies inside the obstacle if $h_j(\mathbf{p}) < 0$, while $h_j(\mathbf{p}) \geq 0$ describes the collision-free region. 
The ground is modeled separately as a horizontal plane at height $z_\mathrm{g}$. 
Including the prescribed safety clearance $d_\mathrm{g}$, the minimum admissible altitude is 
\begin{equation}
    z_{\min} = z_{\mathrm{g}} + d_\mathrm{g}.
\end{equation}
This representation assumes locally level terrain and does not account for ground inclination or uneven terrain.

\subsubsection{Obstacle Constraints and Penalties}
The obstacle conditions extend the \ac{nmpc} formulation at the beginning of this section. 
Together with the \ac{uav} constraints, every ellipsoid selected for the optimization introduces the hard geometric constraint
\begin{equation}
    h_j(\mathbf{p}_k) \geq 0, \qquad k = 1, \ldots, N,
    \label{eq:hard_obs_constraint}
\end{equation}
at the prediction nodes. 
Consequently, a feasible solution cannot place a prediction node inside any selected ellipsoid. 

The hard constraint \eqref{eq:hard_obs_constraint} defines the collision-free region but does not actively encourage the predicted trajectory to maintain a larger separation from the obstacle while remaining feasible. 
Therefore, a soft obstacle penalty is additionally included in the objective function. 
The signed radial distance from the ellipsoid boundary is approximated by
\begin{equation}
    d_{j,k} = \| \mathbf{p}_k - \mathbf{c}_j \|_2 \left( 1- \frac{1}{\sqrt{h_j(\mathbf{p}_k) + 1}} \right). 
\end{equation}
The corresponding bounded obstacle potential is
\begin{equation}
    J_{obs,j,k} = \frac{1}{2} \left[ 1 - \tanh \left( \frac{\alpha_{\mathrm{obs}}}{2} \left( d_{j,k} - \varepsilon_{\mathrm{obs}} \right) \right) \right],
\end{equation}
where $\varepsilon_{\mathrm{obs}}$ determines the distance at which the penalty becomes active and $\alpha_{\mathrm{obs}}$ controls its steepness. 
The soft obstacle penalty at prediction node $k$ is
\begin{equation}
    J_{\mathrm{obs}, k} = \sum_{\mathcal{E}_j \in \mathcal{E}_{\mathrm{act}}} J_{obs,j,k}^2,
    \label{eq:soft_obs_penalty}
\end{equation}
where $\mathcal{E}_{\mathrm{act}} \subseteq \mathcal{E}$ is the set of obstacles included in the current optimization. 
The penalty is small far from an obstacle, and increases smoothly near its boundary, thereby encouraging the optimizer to initiate an avoidance maneuver before the hard constraint becomes active. To account for both the obstacle clearance and the rate at which the \ac{uav} approaches the obstacle, an additional acceleration-level condition is constructed from the first two derivatives of $h_j$ as follows
\begin{align}
    \dot{h}_{j,k} &= 2(\mathbf{p} - \mathbf{c}_j)^\top \mathbf{A}_j \mathbf{v}_k, \\
    \ddot{h}_{j,k} &= 2 \mathbf{v}_k^\top - \mathbf{A}_j \mathbf{v}_k + 2 (\mathbf{p}_k - \mathbf{c}_j)^\top \mathbf{A}_j \mathbf{a}_k,
\end{align}
where $\mathbf{a}_k = \dot{\mathbf{v}}_k$ is the translational acceleration at the prediction node $k$.
Following the \ac{hocbf} \cite{harms2025safe,xiao2022hocbf}, the implemented acceleration-level condition is written directly as 
\begin{equation}
    \ddot{h}_{j,k} + k_\mathrm{vel} \dot{h}_{j,k} + k_\mathrm{pos} h_j(\mathbf{p}_k) \geq -s_{j,k}, \qquad s_{j,k} \geq0, 
\end{equation}
where $k_\mathrm{pos} > 0$ and $k_\mathrm{vel} > 0$ determine the response to the obstacle clearance and approach rate, respectively. 
A newly perceived obstacle may require an acceleration that cannot be produced immediately under the vehicle constraints. 
Therefore, the acceleration-level condition is softened by $s_{j,k}$, which is penalized in the \ac{nmpc} objective. 
The geometric constraint $h_j(\mathbf{p}_k) \geq 0$ remains hard, and no separate first-order \ac{hocbf} constraint is imposed for \ac{lidar} obstacles.

\subsubsection{Obstacle Selection and Trajectory Validation}
The generated solver contains a fixed maximum number $N_{\mathrm{obs}}$ of obstacle slots. 
If $N_e > N_{\mathrm{obs}}$, the obstacles are ranked according to their minimum normalized separation from the remaining prediction
\begin{equation}
    S_j = \min_{\substack{k=0,\ldots,N-1\\\lambda\in[0,1]}} h_j\left((1-\lambda)\mathbf{p}_k + \lambda\mathbf{p}_{k+1} \right).
\end{equation}
The obstacles with the lowest values of $S_j$ form the active set $\mathcal{E}_\mathrm{act}$. 
A trajectory intersecting any ellipsoid or violating the minimum ground altitude is rejected. 
The remaining previous prediction is then checked against the updated obstacle set and reused only if it remains collision-free.

\subsection{Cost Function}
\label{sec:cost_function}
The stage and terminal costs of \eqref{eq:ocp_cost} are formulated as nonlinear least squares,
\begin{equation}
    \label{eq:nls_cost}
    \ell=\tfrac{1}{2}\big\|\costres_k-\costres^{*}_k\big\|^{2}_{\weightmat},
    \qquad
    \ell_e=\tfrac{1}{2}\big\|\costres^{e}-\costres^{e*}\big\|^{2}_{\weightmat^{e}},
\end{equation}
with $\|\bm{e}\|^{2}_{\weightmat}=\bm{e}^\top\weightmat\bm{e}$ and the stage residual
\begin{equation}
    \label{eq:cost_residual}
    \costres_k=\begin{bmatrix}
        \positionvector_k^\top & \bm{\quaternion}_{e,k}^\top & \linearvelvector_k^\top &
        \angularvelvector_k^\top & \thrustvector_k^\top & \uavinput_k^\top &
        \dot{\angularvelvector}_k^\top & J_{\mathrm{obs},k}
    \end{bmatrix}^\top,
\end{equation}
where $\bm{\quaternion}_{e,k}\in\Rthree$ is the vector part of the error quaternion between $\orientation_k$ and the reference attitude.
The corresponding weight is
\begin{equation}
    \label{eq:cost_weight}
    \weightmat=\mathrm{blkdiag}\big(
      \errpenmat,\ \thrustpenmat,\ \inputpenmat,\ \angaccpenmat,\ Q_{\mathrm{obs}}\big).
\end{equation}
The block $\errpenmat$ penalizes the tracking error in position, attitude, linear velocity, and body rate.
The remaining blocks are specific to the thrust-augmented formulation, $\thrustpenmat$ regulates the thrust state $\thrustvector_k$ towards its hover value, $\inputpenmat$ penalizes the control $\uavinput_k=\dot{\thrustvector}_k$ and therefore forbids bang-bang thrust.
$\angaccpenmat$ penalizes the angular acceleration, and $Q_{\mathrm{obs}}$ weights the obstacle penalty $J_{\mathrm{obs},k}$ defined in \autoref{sec:collision_avoidance}.
The terminal residual $\costres^{e}$ drops the two control-dependent entries $\uavinput_k$ and $\dot{\angularvelvector}_k$, with $\weightmat^{e}=\mathrm{blkdiag}(\errpenmat,\thrustpenmat,Q_{\mathrm{obs}})$.
The time steps $\Delta t_k$ are not uniform.
Short steps near the current state preserve integration accuracy, while longer steps at the end of the horizon extend the horizon's reach with a fixed number of shooting nodes.

\subsection{Implementation Details}
\label{sec:implementation_details}
The planner is integrated into the open-source \ac{uav} system of \cite{Baca_2021} and runs at \SI{100}{\hertz}.
% The solver is generated with \texttt{acados} \cite{Verschueren2021}, using a Gauss--Newton Hessian approximation, an explicit Runge--Kutta integrator and a partially condensed \ac{qp} solved by HPIPM.
% A single iteration is performed per call, following the real-time-iteration scheme.
The horizon consists of $\mpchorizon=28$ shooting nodes spanning \SI{2.0}{\second}, distributed as five steps of \SI{10}{\milli\second}, five of \SI{50}{\milli\second}, five of \SI{80}{\milli\second} and thirteen of \SI{100}{\milli\second}.
The number of obstacle slots is fixed at $N_{\mathrm{obs}}=15$ when the solver is generated, so the dimensions of the problem do not change during the flight.
Each iteration is initialized by sampling the stored prediction at the elapsed time.
The optimization and the reference output are decoupled. 
The solver runs on its own timer, while the command passed to the controller is produced at every control step by interpolating the stored prediction between the two nodes surrounding its current age, and the prediction is discarded once it grows older than the horizon, after which the tracker falls back to commanding the current vehicle state until a new solution arrives.

\section{Experiments and Results}
% Values transcribed from tmux/evaluation/simulation/config/custom_config_nmpc_forest.yaml
% (mrs_uav_managers/constraint_manager). Verified identical in custom_config_nmpc.yaml,
% custom_config.yaml and custom_config_csv.yaml, so every arm of every experiment used the
% same definitions. Snap equals jerk in all four profiles and is therefore not listed.
% The ConstraintManager also carries angular_speed and tilt entries, but those are not what
% binds: it declares 60/60/10 rad/s body rates, while the planner's own uav_constraints cap
% them at 6/6/3 rad/s, and the planner's tilt limit is 75 deg against the manager's 60 deg.
% V12 (forest only) is not a single definition, so its cells give the range in effect:
%  - v_xy: the proposed planner received max_linear_speed 12 m/s in four batches and 13 m/s in
%    20260914_194753 (Telemetry acados_params in each bag); SUPER ran 13 m/s.
%  - a_xy, a_z up: not recorded by the planner. NMPC 10 / 11.5 and 4.0 / 5.5 come from each
%    batch's config_used analysis limits, which recorded the tilt change correctly but kept a
%    stale 13 m/s speed, so treat them as the best available record; SUPER 11.5 and 5.5 come
%    from custom_config_super.yaml, whose logged speed and tilt match the runs.
%  - v_z 3 m/s (confirmed in both), a_z down 2 m/s^2, jerk 120 m/s^3 from the configs.
\begin{table}[b]
  \centering
  \scriptsize
  \setlength{\tabcolsep}{4pt}
  \caption{The constraint profiles. The snap limit equals the jerk limit in every profile.}
  \label{tab:constraint_profiles}
  \begin{tabular}{lccccc}
    \toprule
    Profile & $v_{xy}$ & $a_{xy}$ & $j_{xy}$ & $v_{z}$ & $a_{z}$ $\uparrow$/$\downarrow$ \\
     & [\si{\metre\per\second}] & [\si{\metre\per\second\squared}] & [\si{\metre\per\second\cubed}] & [\si{\metre\per\second}] & [\si{\metre\per\second\squared}] \\
    \midrule
    \textit{slow}       & 1.0 & 1.0  & 20  & 1.0 & 1.0/1.0 \\
    \textit{medium}     & 4.0 & 2.0  & 40  & 2.0 & 1.0/1.0 \\
    \textit{fast}       & 8.0 & 4.0  & 60  & 4.0 & 2.0/2.0 \\
    \textit{super fast} & 10.0 & 10.0 & 120 & 3.0 & 3.0/2.0 \\
    \textit{agile}        & 13.0 & 15.0 & 120 & 3.0 & 4.0/2.0 \\
    \bottomrule
  \end{tabular}
  \vspace{-1.5em}
\end{table}

\subsection{Experimental methodology}
This section evaluates \ac{nmpp} in simulation and real-world experiments. 
First, the tracking accuracy of the generated reference is compared with a linear \ac{mpc} trajectory planner \cite{baca2018model} and a polynomial trajectory planner \cite{wang2021large}.
Collision avoidance is then evaluated in a simulated forest. 
Finally, the complete system is deployed in the real \ac{uav} platform.
In every simulated comparison the evaluated planners provide their reference to the same unmodified $SE(3)$ controller~\cite{lee2010geometric}, so that the comparison isolates the reference planner, and each configuration is flown ten times, or fifty times under the \textit{agile} profile. 
All flights are performed under one of the constraint profiles of \autoref{tab:constraint_profiles}, which are enforced by the underlying \ac{uav} system. 
The profiles are identical for all evaluated planners.
The planner's thrust and attitude limits are fixed and do not change with the profile.

% \begin{itemize}
%     \item Without obstacles
%     \begin{itemize}
%         \item Compare with MRS trajectory tracker \cite{baca2018model}
%         \item Compare with the polynomial method \cite{wang2021large}
%     \end{itemize}
%     \item With obstacles
%     \begin{itemize}
%         \item I would like to compare it against \cite{loquercio2021learning, ren2022bubble, zhang2025learning} and \cite{ren2025safety}
%         \item Can I compare it against gmppi? \cite{pochobradsky2026geometric}
%     \end{itemize}
% \end{itemize}

\subsection{Simulation Results}
The simulations use a simulated multirotor in Gazebo with an emulated \texttt{PX4} autopilot, while the trajectory planner and the remaining control components run within the open-source \ac{uav} system \cite{Baca_2021}. 
The same vehicle model and state estimator are used for all evaluated trajectory planners, and the simulations run on a laptop with an Intel Core i7-13700H processor.

\subsubsection{Reference Tracking}
The obstacle-free experiment examines whether including the \ac{uav} dynamics, attitude, and thrust evolution directly in the trajectory planning improves the trackability of the resulting reference. 
% Generated by scripts/make_speed_table.py - do not edit by hand.
% Sources:
%   Slow     nmpc  20260909_123116_nmpc_tracker_se3_slow
%   Slow     mpc   20260909_125152_mpc_tracker_se3_slow
%   Slow     poly  20260909_162439_csv_poly_se3_slow
%   Medium   nmpc  20260909_105003_nmpc_tracker_se3_medium
%   Medium   mpc   20260909_113230_mpc_tracker_se3_medium
%   Medium   poly  20260909_152630_csv_poly_se3_medium
%   Fast     nmpc  20260901_160507_nmpc_tracker_se3_fast
%   Fast     mpc   20260901_164030_mpc_tracker_se3_fast
%   Fast     poly  20260904_110647_csv_poly_se3_fast
%   Super Fast nmpc  20260909_174130_nmpc_tracker_se3_super_fast
%   Super Fast mpc   20260909_175738_mpc_tracker_se3_super_fast
%   Super Fast poly  20260909_182236_csv_poly_se3_super_fast
%   Agile      nmpc  20260912_135915_nmpc_tracker_se3
%   Agile      nmpc  20260912_141423_nmpc_tracker_se3
%   Agile      nmpc  20260912_143039_nmpc_tracker_se3
%   Agile      mpc   20260915_104934_mpc_tracker_se3
%   Agile      poly  20260915_110130_csv_poly_se3
% Requires: \usepackage{booktabs}, \usepackage{siunitx}
\begin{table}[t]
  \centering
  \scriptsize
  \setlength{\tabcolsep}{2pt}
  \caption{Mean 3D \ac{rmse} (\si{\metre}) of a \SI{20}{\metre} flight in free space for varying constraint profiles. $v_{\text{peak}}$ is the range of peak speeds reached across the three planners in (\si{\metre\per\second}) while Reduction denotes the percentage decrease in error relative to each baseline. No standard deviation exceeded \SI{0.045}{\metre}.}
  \label{tab:speed_sweep}
  \begin{tabular}{lccccrr}
    \toprule
    Profile & $v_{\text{peak}}$ & \textbf{NMPP} & \textbf{MPC} & \textbf{Poly} & \textbf{Red.} & \textbf{Red.} \\
     &  & (proposed) & (baseline) & (baseline) & vs MPC & vs Poly \\
    \midrule
    \textit{slow} & 1.0--1.2 & $\mathbf{0.032}$ & $0.077$ & $0.055$ & \textbf{58.2} & \textbf{41.3} \\
    \textit{medium} & 3.9--4.1 & $\mathbf{0.078}$ & $0.238$ & $0.192$ & \textbf{67.3} & \textbf{59.5} \\
    \textit{fast} & 5.9--7.2 & $\mathbf{0.122}$ & $0.373$ & $0.307$ & \textbf{67.4} & \textbf{60.3} \\
    \textit{super fast} & 7.3--8.8 & $\mathbf{0.178}$ & $0.497$ & $0.398$ & \textbf{64.1} & \textbf{55.2} \\
    \textit{agile} & 9.5--10.0 & $\mathbf{0.186}$ & $0.545$ & $0.624$ & \textbf{65.9} & \textbf{70.2} \\
    \bottomrule
  \end{tabular}
  % \vspace{-0.4cm}
  \vspace{-1.5em}
\end{table}

\begin{table}[b]
    \centering
    \scriptsize
    \setlength{\tabcolsep}{3pt}
    \caption{Tracking and computational performance of \ac{nmpp} in the simulated forest. The \ac{rmse} is pooled over the ten flights of each profile. $t_{\mathrm{mean}}$ and $t_{\mathrm{p95}}$ cover the \ac{nmpc} update only.}
    \label{tab:forest_performance}
    \begin{tabular}{lcccc}
    \toprule
    Profile & RMSE [\si{\metre}] & $v_{\max}$ [\si{\metre\per\second}] & $t_{\mathrm{mean}}$ [\si{\milli\second}] & $t_{\mathrm{p95}}$ [\si{\milli\second}] \\
    \midrule
    \textit{slow}       & 0.041 & 1.47 & 5.51 & 7.88 \\
    \textit{medium}     & 0.096 & 4.29 & 5.86 & 8.66 \\
    \textit{fast}       & 0.143 & 7.08 & 5.95 & 9.57 \\
    \textit{super fast} & 0.172 & 9.51 & 5.74 & 8.92 \\
    \textit{agile}        & 0.198 & 11.38 & 6.91 & 9.79 \\
    \bottomrule
    \end{tabular}
    \vspace{-1.5em}
\end{table}

As shown in \autoref{tab:speed_sweep}, the advantage of \ac{nmpp} is retained across all evaluated constraint profiles. 
The worst-case error follows the same trend, averaged over the ten flights of each profile, the maximum tracking error ranges from \SIrange{0.07}{0.55}{\metre} for \ac{nmpp}, against \SIrange{0.12}{1.23}{\metre} for the linear \ac{mpc} trajectory planner and \SIrange{0.10}{1.4}{\metre} for the polynomial trajectory planner.
The absolute gap to the linear \ac{mpc} trajectory planner widens from \SI{0.045}{\metre} under the \textit{slow} profile to \SI{0.359}{\metre} under the \textit{agile} profile.

\subsubsection{Obstacle avoidance}
The second experiment evaluates whether \ac{nmpp} retains its tracking performance when the reference must be modified online according to local \ac{lidar} measurements. 
We use a Poisson forest \cite{karaman2012high}, containing trees with diameter of \SI{0.6}{\meter} and a density of $\delta = \frac{1}{25}~\text{tree} \cdot m^{-2}$. 
This environment follows the forest configuration commonly used for evaluating high-speed \ac{uav} navigation \cite{loquercio2021learning, ren2022bubble, zhang2025learning}.
The simulated \ac{uav} is equipped with an Ouster OS0-128 LiDAR operating at \SI{20}{\hertz}. 
% The planner receives only the simulated sensor measurements, while ground-truth information is used exclusively for the subsequent evaluation. 
Each flight targets a goal \SI{40}{\metre} ahead and counts as successful when the vehicle reached it without a crash, a failsafe or a timeout.

\begin{figure}[t]
    \centering
    \includegraphics[width=\columnwidth]{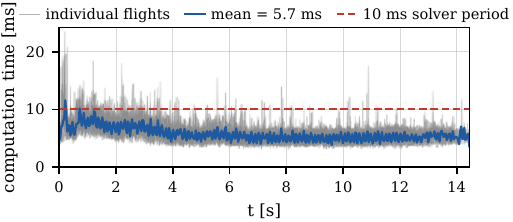}
    \caption{\ac{nmpp} solution time over the ten simulated forest flights under the \textit{super fast} profile. 
    % Grey traces are the individual flights, the solid line is their mean at each instant, and the dashed line marks the \SI{10}{\milli\second} period of the trajectory planner loop.
    % The \SI{10}{\milli\second} update period is not a strict solver timeout. If a new valid solution is unavailable, the trajectory planner continues executing the remaining portion of the previously computed trajectory. A new interpolated trajectory then replaces it only after successful computation and collision validation.
    }
    \label{fig:computation_time}
    \vspace{-1.5em}
\end{figure}

\begin{figure}[b]
    \centering
    \includegraphics[width=1\columnwidth]{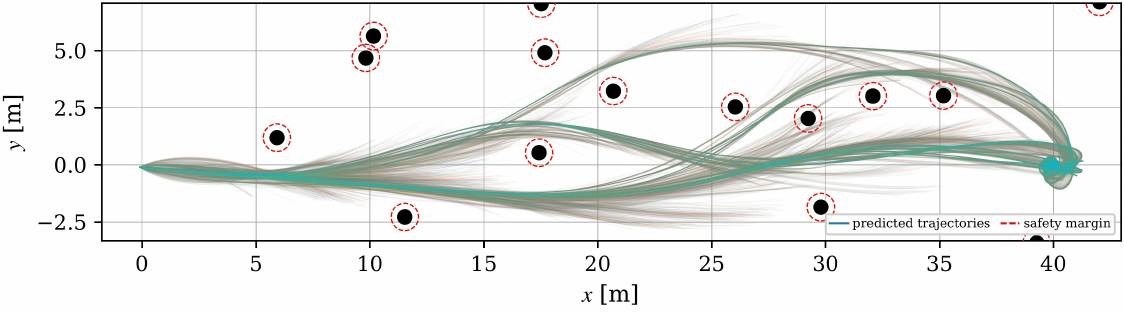}
    \caption{Predicted trajectories obtained during the ten \textit{super fast} simulation runs. Obstacles are shown in black with red-dashed safety margins.}
    \label{fig:predticted_traj}
    \vspace{-1.5em}
\end{figure}

The tracking performance of \ac{nmpp} in the forest experiments is summarized in \autoref{tab:forest_performance}.
While the tracking error increases with trajectory aggressiveness, the mean \ac{nmpc} solution time stays within \SIrange{5.5}{7.0}{\milli\second} across all profiles.
The per-update solution times of the \textit{super fast} profile are shown in \autoref{fig:computation_time}. 
% The mean update takes \SI{5.7}{\milli\second} and the \num{95}th percentile \SI{8.9}{\milli\second}, so a typical update completes within the \SI{10}{\milli\second} period of the \SI{100}{\hertz} trajectory planner loop. 
Although individual solves exceed the \SI{10}{\milli\second} period of the planner loop, mostly within the first second of the flight, the reference is not interrupted, because the command is interpolated from the stored prediction (\autoref{sec:implementation_details}).
The predicted motion over all ten flights with the \textit{super fast} constraint profile is summarized in \autoref{fig:predticted_traj}.
The lines crossing some of the more distant obstacles are predictions computed before the \ac{lidar} sensor could scan them.

\begin{table}[t]
  \centering
  \scriptsize
  \setlength{\tabcolsep}{2pt}
  \caption{Success rate and clearance in the forest. Min.\,clr.\ is the smallest airframe-to-trunk gap over the successful flights, with the propeller tips reaching \SI{0.231}{\metre} from the vehicle centre, so a negative value is a propeller strike the flight survived. The \textit{agile} row is from fifty flights per method.}
  \label{tab:forest_comparison}
  \begin{tabular*}{\columnwidth}{@{\extracolsep{\fill}}lcccc@{}}
    \toprule
    Profile & \multicolumn{2}{c}{NMPP (proposed)} & \multicolumn{2}{c}{SUPER} \\
     & succ.\,[\si{\percent}] & min.\,clr.\,[\si{\metre}] & succ.\,[\si{\percent}] & min.\,clr.\,[\si{\metre}] \\
    \midrule
    \textit{slow} & 100 & +1.243 & 100 & -0.041 \\
    \textit{medium} & 100 & +0.800 & 100 & +0.004 \\
    \textit{fast} & 100 & +0.270 & 100 & +0.183 \\
    \textit{super fast} & 100 & +0.264 & 70 & +0.158 \\
    \textit{agile} & 86 & +0.131 & 26 & -0.145 \\
    \bottomrule
  \end{tabular*}
  \vspace{-1.5em}
\end{table}

To assess relative performance under identical conditions, \ac{nmpp} is compared with SUPER \cite{ren2025safety}. 
Both planners use the same simulated world and provide their references to the same $SE(3)$ controller. 
Their prediction horizons are configured to cover comparable time intervals.
Both methods reached the goal in every run up to the \textit{fast} profile.
Under the \textit{super fast} profile \ac{nmpp} maintains a \SI{100}{\percent} success rate, but the success rate of SUPER decreases to \SI{70}{\percent} (see \autoref{tab:forest_comparison}). 
Under the \textit{agile} profile, \ac{nmpp} reached the goal in 43 of 50 flights and SUPER in 13, with peak speeds of \SI{11.4}{\metre\per\second} and \SI{13.2}{\metre\per\second}, respectively. 
% The trunk clearance of the proposed method stayed positive in all fifty flights, including the seven that did not reach the goal, whereas that of SUPER became negative in nineteen of them. 
% The clearance separates them further: the proposed planner stays at least \SI{0.26}{\metre} from a trunk at every profile, whereas SUPER passes within \SI{0.16}{\metre} of one under the \textit{super fast} profile and its clearance becomes negative under the \textit{slow} profile, where the propeller disc intersected a trunk on a flight that was still counted as successful. 
\subsection{Real-world experiments}
\ac{nmpp} is deployed on a custom-built quadrotor with a take-off mass of \SI{3.25}{\kilogram}, based on a Readytosky Mark4 10-inch V2 frame with a wheelbase of \SI{427}{\milli\metre}.
% The frame is a stretched X configuration, which places the rotors \SI{0.135}{\metre} from the centre of mass along the body $x$ axis and \SI{0.165}{\metre} along the body $y$ axis.
The propulsion consists of four iFlight XING-E 3314 \SI{900}{\kilo\volt} motors with three-blade Gemfan 1045-3 propellers, driven by a Skystars Jupiter 80\,A BL32 electronic speed controller.
The low-level attitude-rate loop runs on a Holybro Pixhawk 6C mini flight controller with the \texttt{PX4} firmware, which is commanded by the same \ac{uav} system~\cite{Baca_2021} running on a Jetson Orin NX with an A603 carrier board.
State estimation fuses the onboard \ac{imu} with a Holybro F9P Helical \ac{rtk} \ac{gnss} receiver and the obstacle measurements are provided by a Livox Mid-360 \ac{lidar}.
% and a RadioMaster XR2 Nano ExpressLRS receiver provides the manual safety link.
% Because the rotor positions differ in the two body axes, the allocation \eqref{eq:motorallocationmat} is used in its general form rather than the symmetric one.

The real-world experiments consist of ten flights through the approximately \SI{40}{\meter} obstacle course under the \textit{fast} constraint profile. 
The experiments were conducted in strong and gusty wind, which disturbed the lightweight pylons as well as the \ac{uav}. 
Consequently, the obstacle boundaries were not perfectly stationary during individual flights, which introduced additional perception uncertainty not present in the simulation. 

\begin{figure}[t]
    \centering
    \includegraphics[width=\columnwidth]{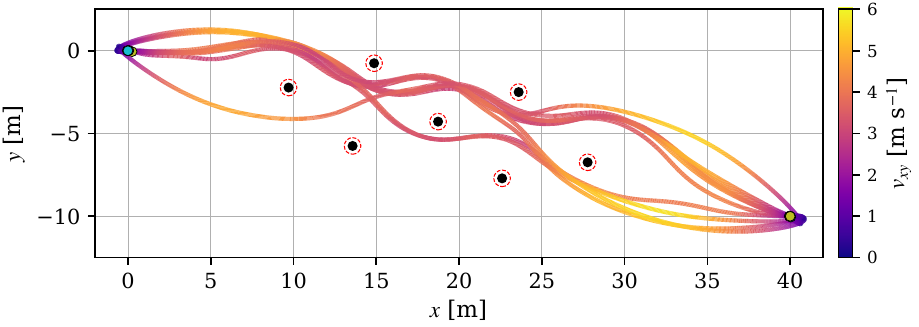}
    \caption{Real \ac{uav} trajectories with corresponding speeds.}
    \label{fig:real_speed}
    \vspace{-1.5em}
\end{figure}

The aggregated real-world results are summarized in \autoref{tab:real_world_performance} and the resulting flown trajectories are shown in \autoref{fig:real_speed}. 
On the onboard computer the mean solution time rises to \SI{13.5}{\milli\second}, above the \SI{10}{\milli\second} period of the planner loop, so a newly optimized trajectory is not available at every control step. 
Because the command is interpolated from the stored prediction (\autoref{sec:implementation_details}), the reference remains continuous and the lower update rate is not visible to the controller.
\begin{table}[t]
    \centering
    \footnotesize
    \setlength{\tabcolsep}{4pt}
    \caption{Aggregated performance over the ten real-world flights under the \textit{fast} profile. The \ac{rmse} is the mean and standard deviation over the individual flights, $v_{max}$ denotes maximum achieved flight speed and $t_{mean}$ and $t_\mathrm{p95}$ denote the mean and 95th-percentile \ac{nmpp} computation times, respectively.}
    \label{tab:real_world_performance}
    \begin{tabular}{ccccc}
        \toprule
        $N_\mathrm{flights}$ & RMSE [\si{\metre}] & $v_{\max}$ [\si{\metre\per\second}] & $t_{\mathrm{mean}}$ [\si{\milli\second}] & $t_{\mathrm{p95}}$ [\si{\milli\second}] \\
        \midrule
        10 & $0.328 \pm 0.070$ & 5.55 & 13.52 & 20.03 \\
        \bottomrule
    \end{tabular}
\end{table}
\section{Conclusion}
In this paper, we introduced \ac{nmpp}, an \ac{nmpc}-based trajectory planner that provides dynamically feasible and collision-free trajectories. 
Because the optimization is formulated over the \ac{uav} model and its actuation limits, the resulting reference can be tracked closely by an unmodified $SE(3)$ controller.
The position \ac{rmse} is reduced by \SIrange{58}{67}{\percent} against a linear \ac{mpc} trajectory planner \cite{baca2018model} and by \SIrange{41}{70}{\percent} against a polynomial trajectory planner \cite{wang2021large} driving the same controller. 
In a Poisson forest, the planner reached the goal without collision in all ten flights of each of the first four constraint profiles, whereas the success rate of SUPER \cite{ren2025safety} decreased to \SI{70}{\percent} under the \textit{super fast} constraint profile.
Under a more aggressive profile (\textit{agile}), the \ac{nmpp} succeeded in \SI{86}{\percent} of fifty flights and SUPER in \SI{26}{\percent}. 
The planner was deployed on a real \ac{uav}, which avoided obstacles at speeds up to \SI{5.5}{\metre\per\second} in gusty wind, running everything on board.

\bibliographystyle{IEEEtran}
\bibliography{ref.bib}

@IEEEtranBSTCTL{BSTcontrol,
  CTLuse_forced_etal       = "yes",
  CTLmax_names_forced_etal = "1",
  CTLnames_show_etal       = "1"
}

@ARTICLE{Teissing_2024_PMM_mintime,
  author={Teissing, Krystof and Novosad, Matej and Penicka, Robert and Saska, Martin},
  journal={IEEE Robotics and Automation Letters}, 
  title={Real-Time Planning of Minimum-Time Trajectories for Agile UAV Flight}, 
  year={2024},
  volume={9},
  number={11},
  pages={10351-10358},
  doi={10.1109/LRA.2024.3471388}}

@INPROCEEDINGS{Burri_2012_plant_inspection,
  author={Burri, Michael and Nikolic, Janosch and Hürzeler, Christoph and Caprari, Gilles and Siegwart, Roland},
  booktitle={2nd International Conference on Applied Robotics for the Power Industry}, 
  title={Aerial service robots for visual inspection of thermal power plant boiler systems}, 
  year={2012},
  pages={70-75},
  }

@InProceedings{Doherty_2007_SAR,
author="Doherty, Patrick
and Rudol, Piotr",
editor="Orgun, Mehmet A.
and Thornton, John",
title="A UAV Search and Rescue Scenario with Human Body Detection and Geolocalization",
booktitle="AI 2007: Advances in Artificial Intelligence",
year="2007",
publisher="Springer Berlin Heidelberg",
address="Berlin, Heidelberg",
pages="1--13",
}

@inproceedings{lee2010geometric,
  title={Geometric tracking control of a quadrotor UAV on SE (3)},
  author={Lee, Taeyoung and Leok, Melvin and McClamroch, N Harris},
  booktitle={49th IEEE conference on decision and control (CDC)},
  pages={5420--5425},
  year={2010},
  organization={IEEE}
}

@Article{Verschueren2021,
  Title                    = {acados -- a modular open-source framework for fast embedded optimal control},
  Author                   = {Robin Verschueren and Gianluca Frison and Dimitris Kouzoupis and Jonathan Frey and Niels van Duijkeren and Andrea Zanelli and Branimir Novoselnik and Thivaharan Albin and Rien Quirynen and Moritz Diehl},
  Journal                  = {Mathematical Programming Computation},
  Year                     = {2021},
}

@article{Baca_2021,
   title={The MRS UAV System: Pushing the Frontiers of Reproducible Research, Real-world Deployment, and Education with Autonomous Unmanned Aerial Vehicles},
   volume={102},
   number={1},
   journal={Journal of Intelligent \& Robotic Systems},
   publisher={Springer Science and Business Media LLC},
   author={Baca, Tomas and Petrlik, Matej and Vrba, Matous and Spurny, Vojtech and Penicka, Robert and Hert, Daniel and Saska, Martin},
   year={2021},
   month=apr }

@INPROCEEDINGS{ren2022bubble,
  author={Ren, Yunfan and Zhu, Fangcheng and Liu, Wenyi and Wang, Zhepei and Lin, Yi and Gao, Fei and Zhang, Fu},
  booktitle={2022 IEEE/RSJ International Conference on Intelligent Robots and Systems (IROS)}, 
  title={Bubble Planner: Planning High-speed Smooth Quadrotor Trajectories using Receding Corridors}, 
  year={2022},
  volume={},
  number={},
  pages={6332-6339},
  doi={10.1109/IROS47612.2022.9981518}}

@article{loquercio2021learning,
  author = {Antonio Loquercio  and Elia Kaufmann  and René Ranftl  and Matthias Müller  and Vladlen Koltun  and Davide Scaramuzza },
  title = {Learning high-speed flight in the wild},
  journal = {Science Robotics},
  volume = {6},
  number = {59},
  pages = {eabg5810},
  year = {2021},
  doi = {10.1126/scirobotics.abg5810},
}

@misc{zhang2025learning,
  title={Back to Newton's Laws: Learning Vision-based Agile Flight via Differentiable Physics}, 
  author={Yuang Zhang and Yu Hu and Yunlong Song and Danping Zou and Weiyao Lin},
  year={2024},
  eprint={2407.10648},
  archivePrefix={arXiv},
  primaryClass={cs.RO},
}

@article{schoels2020ciao,
  author  = {Schoels, Tobias and Rutquist, Per and Palmieri, Luigi
             and Zanelli, Andrea and Arras, Kai O. and Diehl, Moritz},
  title   = {{CIAO}$^\star$: {MPC}-Based Safe Motion Planning in Predictable
             Dynamic Environments},
  journal = {IFAC-PapersOnLine},
  year    = {2020},
  volume  = {53},
  number  = {2},
  pages   = {6555--6562},
  doi     = {10.1016/j.ifacol.2020.12.072}
}

@inproceedings{mellinger2011minimum,
  author    = {Mellinger, Daniel and Kumar, Vijay},
  title     = {Minimum Snap Trajectory Generation and Control for Quadrotors},
  booktitle = {2011 IEEE International Conference on Robotics and Automation},
  pages     = {2520--2525},
  year      = {2011},
  doi       = {10.1109/ICRA.2011.5980409}
}

@inproceedings{wang2021large,
  title     = {Generating Large-Scale Trajectories Efficiently using Double Descriptions of Polynomials},
  author    = {Wang, Zhepei and Ye, Hongkai and Xu, Chao and Gao, Fei},
  booktitle = {2021 IEEE International Conference on Robotics and Automation (ICRA)},
  pages     = {7436--7442},
  year      = {2021},
  doi       = {10.1109/ICRA48506.2021.9561585}
}

@article{pochobradsky2026geometric,
  title={Geometric model predictive path integral for agile uav control with online collision avoidance},
  author={Pochobradsk{\'y}, Pavel and Proch{\'a}zka, Ond{\v{r}}ej and P{\v{e}}ni{\v{c}}ka, Robert and Von{\'a}sek, Vojt{\v{e}}ch and Saska, Martin},
  journal={IEEE Robotics and Automation Letters},
  year={2026},
  publisher={IEEE}
}

@inproceedings{karaman2012high,
  title={High-speed flight in an ergodic forest},
  author={Karaman, Sertac and Frazzoli, Emilio},
  booktitle={2012 IEEE International Conference on Robotics and Automation},
  pages={2899--2906},
  year={2012},
  organization={IEEE}
}

@inproceedings{rusu2011pcl,
    author    = {Radu Bogdan Rusu and Steve Cousins},
    title     = {{3D is Here: Point Cloud Library (PCL)}},
    booktitle = {IEEE International Conference on Robotics and Automation},
    year      = {2011},
    pages     = {1--4},
    doi       = {10.1109/ICRA.2011.5980567}
  }

@article{xiao2022hocbf,
    author  = {Wei Xiao and Calin Belta},
    title   = {High-Order Control Barrier Functions},
    journal = {IEEE Transactions on Automatic Control},
    volume  = {67},
    number  = {7},
    pages   = {3655--3662},
    year    = {2022}
  }

@inproceedings{khatib1985realtime,
    author    = {Khatib, Oussama},
    title     = {Real-Time Obstacle Avoidance for Manipulators and Mobile Robots},
    booktitle = {Proceedings of the 1985 IEEE International Conference on
                 Robotics and Automation},
    year      = {1985},
    volume    = {2},
    pages     = {500--505},
    doi       = {10.1109/ROBOT.1985.1087247}
  }

@article{fiorini1998motion,
    author  = {Fiorini, Paolo and Shiller, Zvi},
    title   = {Motion Planning in Dynamic Environments Using Velocity Obstacles},
    journal = {The International Journal of Robotics Research},
    year    = {1998},
    volume  = {17},
    number  = {7},
    pages   = {760--772},
    doi     = {10.1177/027836499801700706}
  }

@incollection{vanDenBerg2011reciprocal,
    author    = {van den Berg, Jur and Guy, Stephen J. and Lin, Ming
                 and Manocha, Dinesh},
    title     = {Reciprocal n-Body Collision Avoidance},
    booktitle = {Robotics Research: The 14th International Symposium ISRR},
    series    = {Springer Tracts in Advanced Robotics},
    volume    = {70},
    publisher = {Springer},
    year      = {2011},
    pages     = {3--19},
    doi       = {10.1007/978-3-642-19457-3_1}
  }

@article{karaman2011sampling,
    author  = {Karaman, Sertac and Frazzoli, Emilio},
    title   = {Sampling-Based Algorithms for Optimal Motion Planning},
    journal = {The International Journal of Robotics Research},
    year    = {2011},
    volume  = {30},
    number  = {7},
    pages   = {846--894},
    doi     = {10.1177/0278364911406761}
  }

@incollection{richter2016polynomial,
    author    = {Richter, Charles and Bry, Adam and Roy, Nicholas},
    title     = {Polynomial Trajectory Planning for Aggressive Quadrotor Flight
                 in Dense Indoor Environments},
    booktitle = {Robotics Research: The 16th International Symposium ISRR},
    series    = {Springer Tracts in Advanced Robotics},
    volume    = {114},
    publisher = {Springer},
    year      = {2016},
    pages     = {649--666},
    doi       = {10.1007/978-3-319-28872-7_37}
  }

@article{wang2022geometrically,
    author  = {Wang, Zhepei and Zhou, Xin and Xu, Chao and Gao, Fei},
    title   = {Geometrically Constrained Trajectory Optimization for
               Multicopters},
    journal = {IEEE Transactions on Robotics},
    year    = {2022},
    volume  = {38},
    number  = {5},
    pages   = {3259--3278},
    doi     = {10.1109/TRO.2022.3160022}
  }

@inproceedings{baca2018model,
    author    = {Baca, Tomas and Hert, Daniel and Loianno, Giuseppe
                 and Saska, Martin and Kumar, Vijay},
    title     = {Model Predictive Trajectory Tracking and Collision Avoidance
                 for Reliable Outdoor Deployment of Unmanned Aerial Vehicles},
    booktitle = {2018 IEEE/RSJ International Conference on Intelligent Robots
                 and Systems (IROS)},
    year      = {2018},
    pages     = {6753--6760},
    doi       = {10.1109/IROS.2018.8594266}
  }

@inproceedings{schoels2020nmpc,
    author    = {Schoels, Tobias and Palmieri, Luigi and Arras, Kai O.
                 and Diehl, Moritz},
    title     = {An {NMPC} Approach Using Convex Inner Approximations for
                 Online Motion Planning with Guaranteed Collision Avoidance},
    booktitle = {2020 IEEE International Conference on Robotics and Automation
                 (ICRA)},
    year      = {2020},
    pages     = {3574--3580},
    doi       = {10.1109/ICRA40945.2020.9197206}
  }

@inproceedings{florence2018nanomap,
    author    = {Florence, Peter R. and Carter, John and Ware, Jake
                 and Tedrake, Russ},
    title     = {{NanoMap}: Fast, Uncertainty-Aware Proximity Queries with
                 Lazy Search over Local {3D} Data},
    booktitle = {2018 IEEE International Conference on Robotics and Automation
                 (ICRA)},
    year      = {2018},
    pages     = {7631--7638},
    doi       = {10.1109/ICRA.2018.8463195}
  }

@article{zhou2021egoplanner,
    author  = {Zhou, Xin and Wang, Zhepei and Ye, Hongkai and Xu, Chao
               and Gao, Fei},
    title   = {{EGO-Planner}: An {ESDF}-Free Gradient-Based Local Planner
               for Quadrotors},
    journal = {IEEE Robotics and Automation Letters},
    year    = {2021},
    volume  = {6},
    number  = {2},
    pages   = {478--485},
    doi     = {10.1109/LRA.2020.3047728}
  }

@article{ren2025safety,
    author  = {Ren, Yunfan and Zhu, Fangcheng and Lu, Guozheng and Cai, Yixi
               and Yin, Longji and Kong, Fanze and Lin, Jiarong and Chen, Nan
               and Zhang, Fu},
    title   = {Safety-Assured High-Speed Navigation for {MAVs}},
    journal = {Science Robotics},
    year    = {2025},
    volume  = {10},
    number  = {98},
    pages   = {eado6187},
    doi     = {10.1126/scirobotics.ado6187}
  }

@inproceedings{zhang2025mapless,
    author    = {Zhang, Linzuo and Hu, Yu and Deng, Yang and Yu, Feng
                 and Zou, Danping},
    title     = {Mapless Collision-Free Flight via {MPC} Using Dual
                 {KD}-Trees in Cluttered Environments},
    booktitle = {2025 IEEE/RSJ International Conference on Intelligent Robots
                 and Systems (IROS)},
    year      = {2025},
    pages     = {19653--19660},
    doi       = {10.1109/IROS60139.2025.11247753}
  }

@inproceedings{liang2025pointcloud,
    author    = {Liang, Faduo and Yang, Yunfeng and Dai, Shi-Lu},
    title     = {Point Cloud-Based Control Barrier Functions for Model
                 Predictive Control in Safety-Critical Navigation of
                 Autonomous Mobile Robots},
    booktitle = {2025 IEEE/RSJ International Conference on Intelligent Robots
                 and Systems (IROS)},
    year      = {2025},
    pages     = {2060--2067},
    doi       = {10.1109/IROS60139.2025.11247044}
  }

@inproceedings{harms2025safe,
    author    = {Harms, Marvin and Jacquet, Martin and Alexis, Kostas},
    title     = {Safe Quadrotor Navigation Using Composite Control Barrier
                 Functions},
    booktitle = {2025 IEEE International Conference on Robotics and Automation
                 (ICRA)},
    year      = {2025},
    pages     = {6343--6349},
    doi       = {10.1109/ICRA55743.2025.11127368}
  }

@inproceedings{gupta2025lol,
  title={Lol-nmpc: Low-level dynamics integration in nonlinear model predictive control for unmanned aerial vehicles},
  author={Gupta, Parakh M and Proch{\'a}zka, Ond{\v{r}}ej and H{\v{r}}ebec, Jan and Novosad, Matej and P{\v{e}}ni{\v{c}}ka, Robert and Saska, Martin},
  booktitle={2025 IEEE/RSJ International Conference on Intelligent Robots and Systems (IROS)},
  pages={1186--1193},
  year={2025},
  organization={IEEE}
}

@article{kratky2025rvc,
  title={RVC-NMPC: Nonlinear Model Predictive Control with Reciprocal Velocity Constraints for Mutual Collision Avoidance in Agile UAV Flight},
  author={Kratky, Vit and Penicka, Robert and Gupta, Parakh M and Proch{\'a}zka, Ondrej and Saska, Martin},
  journal={arXiv preprint arXiv:2512.08574},
  year={2025}
}

\end{document}